\documentclass[lettersize,journal]{IEEEtran}
\usepackage{amsmath,amsfonts}
\usepackage{algorithmic}
\usepackage{algorithm}
\usepackage{array}
\usepackage[caption=false,font=normalsize,labelfont=sf,textfont=sf]{subfig}
\usepackage{textcomp}
\usepackage{stfloats}
\usepackage{url}
\usepackage{verbatim}
\usepackage{graphicx}
\usepackage{cite}
\begin{document}

\title{HIFI-Net: A Novel Network for Enhancement for Underwater Images}

\author{Jiajia Zhou, Junbin Zhuang, Yan Zheng*, Di Wu~\IEEEmembership{}
        % <-this % stops a space
\thanks{This paper was supported by NSFC 62101156.}% <-this % stops a space
\thanks{Manuscript received June 4, 2022. Yan Zheng is the corresponding author of this paper.  Emails of authors are zhoujiajia@hrbeu.edu.cn;916800391@qq.com;yanzheng3@hrbeu.edu.cn;
dwu.robotics@gmail.com. They are all in College of Intelligent Systems Science and Engineering, Harbin Engineering University, Harbin 150001, China. }}


\IEEEpubid{}

\maketitle

\begin{abstract}
A novel network for enhancement to underwater images is proposed in this paper. It contains a Reinforcement Fusion Module for Haar wavelet images (RFM-Haar) based on Reinforcement Fusion Unit (RFU), which is used to fuse an original image and some important information within it. Fusion is achieved for better enhancement. As this network make ``Haar Images into Fusion Images", it is called HIFI-Net. The experimental results show the proposed HIFI-Net performs best among many state-of-the-art methods on three datasets at three normal metrics and a new metric.
\end{abstract}

\iffalse
\section{Introduction}
\IEEEPARstart{T}{his} file is intended to serve as a ``sample article file''
for IEEE journal papers produced under \LaTeX\ using
IEEEtran.cls version 1.8b and later. The most common elements are covered in the simplified and updated instructions in ``New\_IEEEtran\_how-to.pdf''. For less common elements you can refer back to the original ``IEEEtran\_HOWTO.pdf''. It is assumed that the reader has a basic working knowledge of \LaTeX. Those who are new to \LaTeX \ are encouraged to read Tobias Oetiker's ``The Not So Short Introduction to \LaTeX ,'' available at: \url{http://tug.ctan.org/info/lshort/english/lshort.pdf} which provides an overview of working with \LaTeX.

\section{The Design, Intent, and \\ Limitations of the Templates}
The templates are intended to {\bf{approximate the final look and page length of the articles/papers}}. {\bf{They are NOT intended to be the final produced work that is displayed in print or on IEEEXplore\textsuperscript{\textregistered}}}. They will help to give the authors an approximation of the number of pages that will be in the final version. The structure of the \LaTeX\ files, as designed, enable easy conversion to XML for the composition systems used by the IEEE. The XML files are used to produce the final print/IEEEXplore pdf and then converted to HTML for IEEEXplore.
\fi
\section{Method}
The Reinforcement Fusion Unit (RFU) is introduced in this section firstly. RFU is used to fuse information in two different kinds of images, in which one is an image containing base information, and the other is an image containing the additional reinforcement information. RFU generates the fusion image (or called fusion feature map), as shown in Fig.1. Then, Reinforcement Fusion Module for Haar (RFM-Haar), which fuses Haar images as reinforcement information into the original underwater image as base information, is introduced. As shown in Fig.2, RFM-Haar is based on RFU. Finally, a novel underwater image enhancement network, which uses RFM-Haar, is proposed as shown in Fig.3. As this network fuses Haar images into a fusion image, we call it HIFI-Net.

\begin{figure}[!t]
\centering
\includegraphics[width=1\columnwidth,height =1\columnwidth]{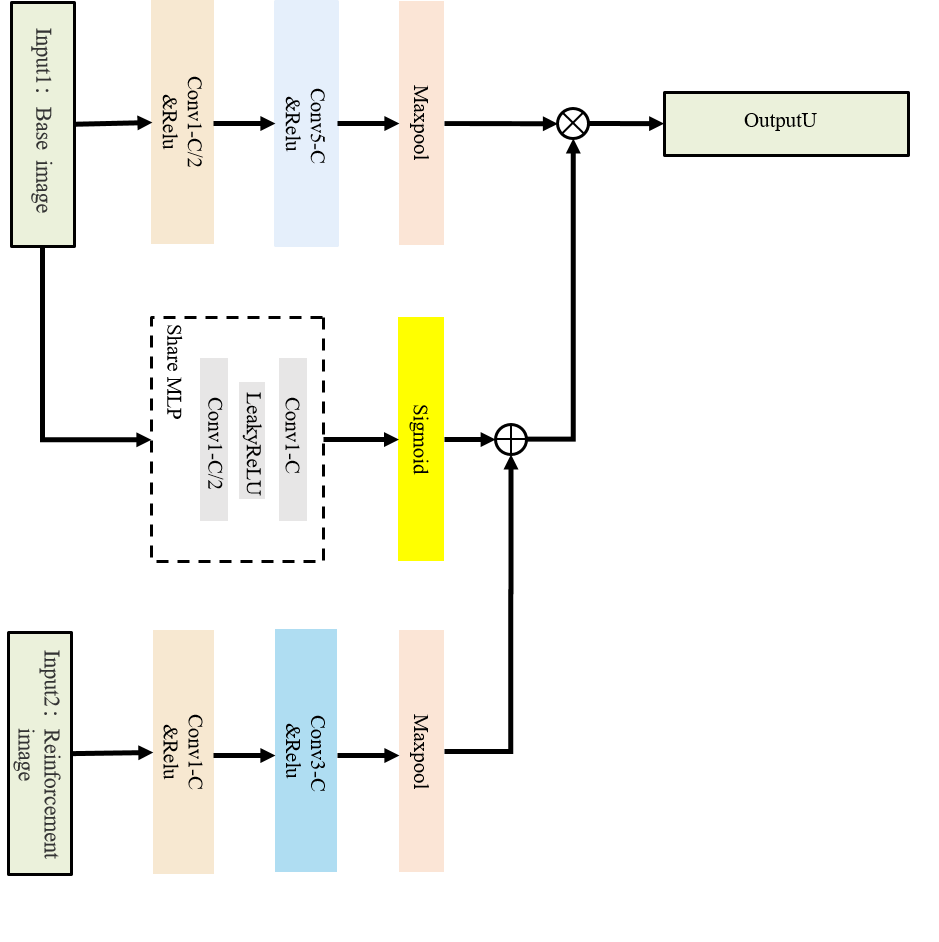}
\centering
\caption{Reinforcement Fusion Unit.}
\label{fig_1}
\end{figure}
 
\subsection{Reinforcement Fusion Unit}
RFU can fuse a base image and a reinforcement image into a new image (which can also be seen as a 3D matrix), as shown in Fig. 1. RFU is universal for images containing different kinds of reinforcement information. Usually, the coordinates of the pixels must have the same physical meaning in a base image and its reinforcement image, in order to achieve ideal fusion. 

The RFU has two inputs ($Input1$ and $Input2$) and one output ($OutputU$). $Input1$ is an image containing base information, called base image. $Input2$ is an image containing additional reinforcement information, called reinforcement image. $OutputU$ is a fusion image. At first, RFU processes $Input1$ using $C/2$ kernels of size $\mathit{1\times 1}$ and $C$ kernels of size $\mathit{5\times 5}$ for convolution. $C$ is the wanted number of channels of the output image. It can be set according to the number of channels of the input image or other reasons. In RFM-Haar, $C$ is $16$ and the stride size of convolutions is always 1. This operation is shown in Eq\eqref{Eq1}.
\begin{align}
\label{Eq1}
\mathit{I_{b}=Maxpool(\phi_{5-c}^{} (\phi_{1-c/2}^{}(Input1)))}
\end{align}
where $\mathit{\phi_{k-c}^{}}$ is the abstract function of convolution processing to an image with $c$ kernels of size $k$ (size $k$ means $\mathit{k\times k}$), and followed by the Relu activation function. Similarly, $Maxpool$ denotes the max-pooling with a pooling kernel of $3$ and a stride size of $1$. The pooling parameters are the same in this paper. 

On the other hand, RFU uses a Shared Multilayer Perceptron(MLP) to extract the features of $Input1$. The Shared MLP is composed of a $\mathit{1\times 1}$ convolution, a ReLU activation function and a $\mathit{1\times 1}$ convolution, as shown in Eq\eqref{Eq2}.
\begin{align}
\label{Eq2}
\mathit{MLP( Input1)=\phi_{1-c}^{'' }(ReLU(\phi_{1-c/2}^{''}(Input1 ))}  
\end{align}
where $\mathit{MLP}$ is the abstract function of the Shared MLP. A Shared MLP adjusts and aggregates the underlying information on a base image.$\mathit{\phi_{k-c}^{''}}$ is convolution operation without activate function.
Then Sigmoid function filters the candidate features, as shown in  Eq\eqref{Eq3}.
\begin{align}
\label{Eq3}
\mathit{M_b=\sigma(MLP(Input1))}
\end{align}
where $\mathit{\sigma}$ denotes the Sigmoid activation function. $\mathit{M_b}$ can be viewed as a matrix of to base information.

$Input2$ is a reinforcement image of size $\mathit{M \times N}$. RFU processes the image using $\mathit{1\times 1}$ and $\mathit{3\times 3}$ convolution kernels. Numbers of kernels in two sizes are all set to $C$, as shown in Eq\eqref{Eq4}.
\begin{align}
\label{Eq4}
\mathit{M_a=Maxpool\left(\phi_{3-c}^{}(\phi_{1-c}^{}(Input2))\right)} 
\end{align}
where max-pooling follows convolutions. The computation result $\mathit{M_a}$ can be viewed as the matrix of coefficients to reinforcement image.

The matrix $M_a$ coming from $Input2$ and the matrix $M_b$ coming from $Input1$ are summed. The fusion intensity control matrix $M_f$ is shown in Eq\eqref{Eq5}.
\begin{align}
\label{Eq5}
\mathit{M_{f}^{}=M_{a}^{}\oplus M_{b}^{}} 
\end{align}
where $\mathit{\oplus}$ denotes each pair of corresponding elements in $M_a$ and $M_b$ are summed to form $M_f$.

Next, $M_f$ is used to adjust $I_b$ and generate the fusion feature map. as show in Eq\eqref{Eq6}.
\begin{align}
\label{Eq6}
\mathit{OutputU =I_b{\otimes} M_f}
\end{align}
where $\mathit{\otimes}$ means that each element in $I_b$ is multiplied by its corresponding element in $M_f$ to form $OutputU$.

\begin{figure*}[!t]
\centering
\includegraphics[width=2.0\columnwidth,height =0.8\columnwidth]{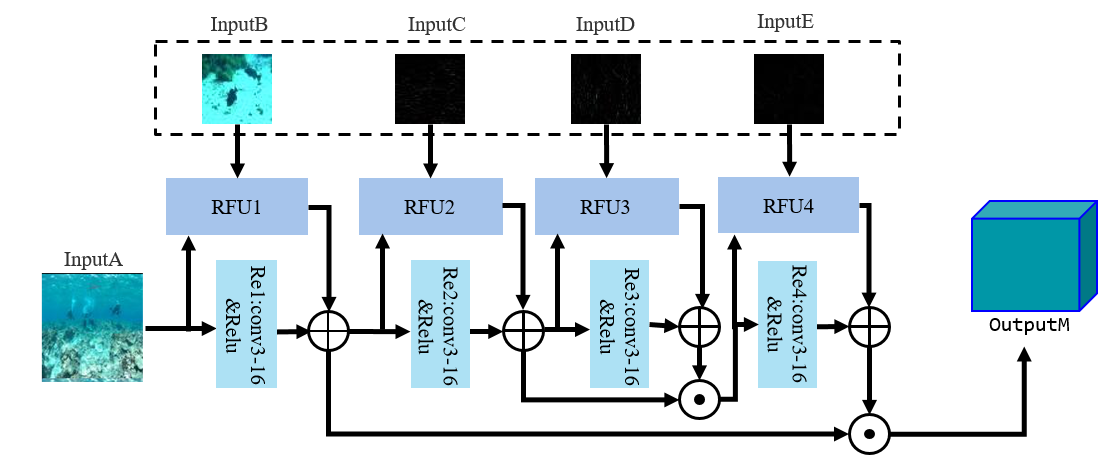}
\centering
\caption{Reinforcement Fusion Module for Haar.}

\label{Fig2}
\end{figure*}
\subsection{Reinforcement Fusion Module for Haar}
RFM-Haar is designed based on RFU. It contains four RFUs and four residual convolution (RE) operations, as shown in Fig.2. The first-order two-dimensional Haar wavelet transforms the original image of size $M \times N$ into four sub-images of size $M/2 \times N/2$. These four sub-images after proper upsampling to size $M \times N$ are fused with a base image to finally generate the fusion image with Haar reinforcement in RFM-Haar. The Haar wavelet transform result after proper upsampling and the original input image meet the spatial correspondence as a condition of RFU usage.

This module for fusing images in four steps is designed for Haar, as it has four sub-images. This module is equally applicable to other image fusion containing five inputs (one base and four reinforcement images). Even for fusing different numbers of images, RFM-Haar is only need to change the hyperparameters of the module appropriately. Therefore, though RFM-Haar is designed for Haar images fusion, the module can also be understood as RFM-RI (RI means reinforcement information). This may be a new way to combine CNN with traditional image processing.

RFM-Haar has five inputs ($InputA$, $InputB$, $InputC$, $InputD$ and $InputE$ ), four RFUs (RFU1, RFU2, RFU3 and RFU4) and four residual blocks (RE1, RE2, RE3 and RE4 ).

$InputA$: An original underwater image ($3$ channels).

$InputB$: The upsampled zero-order image of Haar wavelet transformed $InputA$.

$InputC$: The upsampled first-order x-direction image (horizontal component) of Haar wavelet transformed $InputA$.

$InputD$: The upsampled first-order y-direction image (vertical component) of Haar wavelet transformed $InputA$.

$InputE$: The upsampled first-order xy-direction image (diagonal component) of Haar wavelet transformed $InputA$.

It can be seen that the $InputA$ is pre-decomposed into $4$ sub-images using Haar wavelet. Then, these $4$ sub-images are upsampled to size $M \times N$ using the nearest neighbor interpolation method, as the size of $InputA$ is $M \times N$.

%%%%%%%%%%%%%%%%%%%
%%%%%%%%%%%%%%%%%%%
$InputA$ is firstly used as $Input1$ of RFU1 to participate in the first fusion, as shown in Fig.2. $InputB$ is used as $Input2$ of RFU1, and the $OutputU$ of RFU1 is a fusion image of $M \times N \times 16$. On the other hand, $InputA$ also enters the residual convolution with $16$ kernels of size $3 \times 3$. Here the residual convolution is to solve the network degradation problem.

$InputA$ is used as the input of the first residual block (RE1). The output of RE1 is a feature map of $M \times N \times 16$. This feature map is summed with the $OutputU$ of RFU1 to generate $Input1$ of RFU2. The calculation process from $InputA$, $InputB$ to the $Input1$ of RFU2 is shown in Eqs(7,8).
\begin{align}
\label{Eq7}
\mathit{RFU1.OutputU =F_{RFU}(InputA,InputB ) } \\
\mathit{RFU2.Input1 = RFU1.OutputU \oplus \phi^{\prime}_{3-16}(InputA)}
\end{align}
where, $F_{ RFU }(\cdot ,\cdot )$ is the abstract function of RFU. $RFUj.Input1$, $RFUj.Input2$, $RFUj.OutputU$ represent two inputs and one output of the $j$th RFU respectively, and $j$ can be $1$,$2$,$3$ and $4$. $\phi^{\prime}_{3-16}$ means residual convolution of $3\times3$ kernels.

As Fig.2 shows, the inputs of RFU2 and RFU3 are the outputs of previous stages, $InputC$ and $InputD$.The $RFU4.Input1$ is a spliced feature map in the channel dimension from ${RFU3.Input1}$ and $\mathit{RFU3.OutputU{\oplus} \phi^{\prime}_{3-16}(RFU3.Input1)}$. $ RFU2.OutputU$ and $\mathit{RFU2.Input1 \oplus \phi^{\prime}_{3-16}(RFU4.Input1)} $is spliced to form the output of RFM-Haar.The specific operation is shown in Eq (9-14).
\begin{align}
\label{Eq9}
\mathit{RFU2.OutputU =F_{RFU}(RFU2.Input1, InputC) } \\[1mm]
\mathit{RFU3.Input1 =RFU2.OutputU {\oplus} \phi^{\prime}_{3-16}(RFU2.Input1)} \\[1mm]
\mathit{RFU3.OutputU =F_{RFU} (RFU3.Input1, InputD)} \\
\mathit{RFU4.Input1 =RFU3.Input1 \odot (RFU3.OutputU} \nonumber\\
\mathit{{\oplus} \phi^{\prime}_{3-16}(RFU3.Input1))} \\
\mathit{RFU4.OutputU =F_{RFU} (RFU4.Input1, InputE) } \\
\mathit{OutputM =RFU2.Input1 \odot (RFU4.Output \oplus }\nonumber\\
\mathit{\phi^{\prime}_{3-16} ( RFU4.Input1))}
\end{align}
where $\odot$ denotes channel-wise splicing and $OutputM$ is the output of RFM-Haar.

\begin{figure*}[!t]
\centering
\includegraphics[width=2.0\columnwidth,height =0.7\columnwidth]{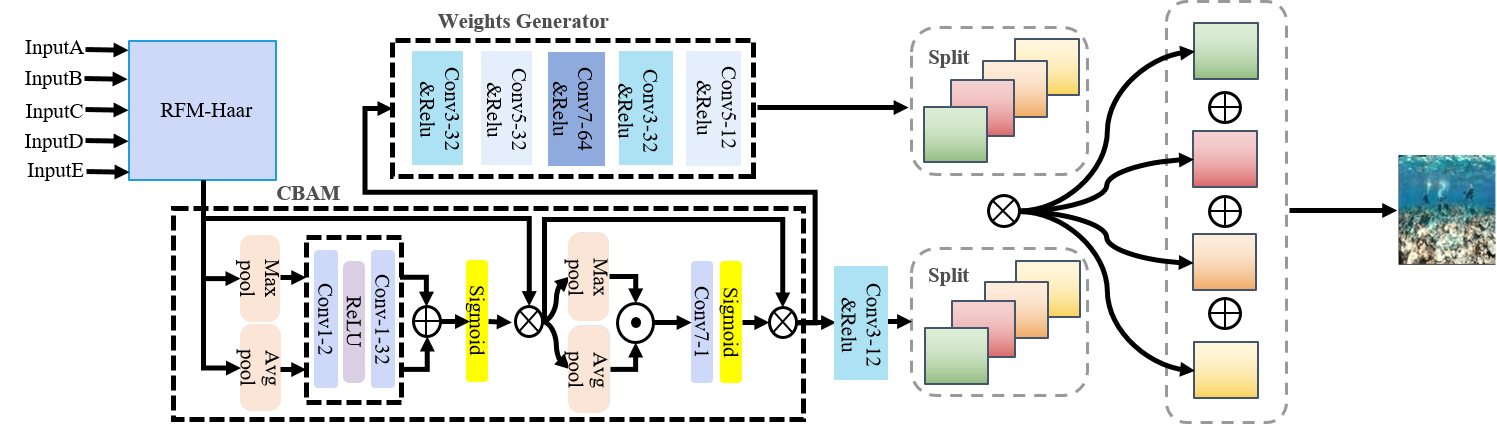}
\centering
\caption{HIFI-Net.}
\label{Fig3}
\end{figure*}

\subsection{HIFI-Net}
We proposed an underwater image enhancement network that uses the RFM-Haar. This network makes ``Haar Images into Fusion Image", so we call it HIFI-Net. As Fig.3 shows, this network has five inputs, the same as RFM-Haar. These five inputs are fused into a new $32$-channel feature image in RFM-Haar. This new image is subsequently processed by the CBAM.

CBAM\cite{woo2018cbam} is composed of a channel attention unit and a spatial attention unit.
\begin{align}
\label{Eq15}
\mathit{ F_1=F_0 \otimes  \sigma (\phi^{''}_{1-32} (ReLU(\phi^{''}_{ 1-2}(Avgpool(F_0))))}\nonumber\\ 
\mathit{\oplus \phi^{''}_{1-32} (ReLU(\phi^{''}_{1-2}(Maxpool(F_0)))))} 
\end{align}
where $F_0$ is the output of RFM-Haar, $F_1$ is the channel attention output; $AvgPool$ is the average pooling operation, kernel size 3, stride size 1.
\begin{align}
\label{Eq16}
\mathit{F_1^{\prime}  =\sigma (\phi_{7-1}^{''} (AvgPool(F_1) \odot MaxPool(F_1))} \\
 \mathit{F_2  =F_1 \otimes F_1^{\prime}}\quad\quad\quad\quad\quad\quad
\end{align}
where $F_{2}$ is the spatial attention output.  In Eq(17), 1-channel image $F_1^{\prime}$ is multiplied by $c^{\prime}$-channels image $F_1$. In this situation, 1-channel image $F_1^{\prime}$ is copied to form a $c^{\prime}$-channels image $I_m$, whose each channel is as the same as $\mathit{F_1^{\prime}}$, then $\mathit{F_{1}\otimes F_{1}^{\prime} = F_{1} \otimes  I_m}$ is established.

$F_{2}$ is convolved by $\mathit{\phi_{3-12}^{}}$. The 12-channel feature map $F_3$ is output, as shown in Eq\eqref{Eq18}.
\begin{align}
\label{Eq18}
\mathit{F_{3}=\phi_{3-12}^{}\left(F_{2}\right)}
\end{align}
%%%%%%%%%%%%%%%%%%%%%%%%%%%%%%%%%%
%%%%%%%%%%%%%%%%%%%%%%%%%%%%%%%%%%%
The output of the weight generator is a feature map of $M \times N \times 12$. It is mainly composed of a series of convolutions of different scales. The details of the operation are shown in Eq\eqref{Eq19}.
\begin{equation}
\begin{split}
\label{Eq19}
\hspace{6mm}
\mathit{W_{all}^{}=\phi_{5-12}^{} (\phi_{3-32}^{} (\phi_{7-64}^{}(\phi_{5-32}^{} (   \phi_{3-32}^{} (F_2)))))} 
\end{split}
\end{equation}
where $F_2$ denotes the output of the CBAM; $\mathit{W_{all}^{}}$ is the output of the weight generator.

Underwater images processed by CNN are prone to contain color casts and artifacts. Therefore, a gated fusion module is used to solve this problem. The gated fusion module uses a weight generator to learn the confidence mapping and splits it into four $M \times N \times 3$ confidence matrices. $F_3$ is also split into four $M \times N \times 3$ feature maps. The importance of these four feature maps is determined using the confidence matrices. Finally, these four feature maps are pixel-wise summed. The operation is shown in Eq\eqref{Eq020}.
\begin{align}
\label{Eq020}
\mathit{I_{en}^{}=(R _{c1}^{}\otimes  C_{w1}^{})  \oplus (R _{c2}^{}\otimes C_{w2}^{}) \oplus}  \nonumber\\ 
\mathit{(R _{c3}^{}\otimes C_{w3}^{})   \oplus (R _{c4}^{}\otimes  C_{w4}^{})  }
\end{align}
where $\mathit{I_{en}^{}}$ is the underwater enhanced image, i.e., the output of HIFI-Net. $R _{c1}^{}$, $R _{c2}^{}$, $R _{c3}^{}$, and $R _{c4}^{}$ are four feature maps. $C_{w1}^{}$, $C_{w2}^{}$, $C_{w3}^{}$, and $C_{w4}^{}$ are confidence matrices.
%%%%%%%%%%%%%%%%%%%%%%%%%%%%%%%
\begin{table*}[]
\centering
    \caption{Full reference image quality was evaluated with Mse on the EUVP, UFO-120 and UIEB test sets.}
\begin{tabular}{cccccccccc}
\hline
\multicolumn{1}{l}{} & \multicolumn{1}{l}{} & \multicolumn{1}{l}{} & \multicolumn{1}{l}{} & \multicolumn{1}{l}{} & \multicolumn{1}{l}{} & \multicolumn{1}{l}{} & \multicolumn{1}{l}{} & \multicolumn{1}{l}{} & \multicolumn{1}{l}{} \\
Mtriy & Datesets & Cycle-GAN\cite{li2018emerging} & DCP\cite{he2010single} & Mul-Fusion\cite{mohan2020underwater} & Water-Net\cite{li2019underwater} & HE\cite{pizer1987adaptive} & FUnIE-GAN\cite{islam2020fast} & UWNet\cite{naik2021shallow} & Ours \\
 &  &  &  &  &  &  &  &  &  \\ \hline
 &  &  &  &  &  &  &  &  &  \\
 & EUVP & 0.6173 & 0.8591 & 2.2212 & 0.2663 & 1.9295 & 0.2966 & 0.3593 & \textbf{0.1952} \\
MSE $\downarrow$ &  &  &  &  &  &  &  &  &  \\
$ ( \times 10_{}^{3} )$ & UFO-120 & 0.4624 & 0.7217 & 2.4803 & 0.2358 & 1.9080 & 0.2639 & 0.2596 & \textbf{0.1512}\\
 &  &  &  &  &  &  &  &  &  \\
 & UIEB &0.7496 & 1.1419 & 1.8620 & 0.7975 & 1.1610 & 1.8135 & 0.8766 & \textbf{0.4683}\\
 &  &  &  &  &  &  &  &  &  \\ \hline
\end{tabular}
\end{table*}
%%%%%%%%%%%%%%%%%%%%%%%%%%%%%%%%%%%%%%%%
\subsection{Objective function}
To train HIFI-Net, the loss function is divided into three parts, which are $L_{cha}^{}$, $L_{ssim}^{}$ and $L_{per}^{}$. The first part need to make the whole network sample and learn from the global similarity space. The robust Charbonnier loss\cite{lai2018fast} function is $L_{cha}^{}$. It mainly focuses on the overall feature similarity of the image content, which is described in Eq\eqref{Eq021}.
\begin{align}
\label{Eq021}
\mathit{L_{cha}\left(J_{c}, \hat{J}_{c}\right)=\sqrt{\left(J_{c}-\hat{J}_{c}\right)^{2}+\varepsilon^{2}}}
\end{align}
where $\hat{J}_{c}$ and $J_{c}$ are the ground truth and enhanced image respectively, and $\varepsilon$ is a very small constant.

$L_{ssim}$ is the similarity index (SSIM) loss function\cite{wang2004image}. It improves the local structure and details. SSIM can be defined as.
\begin{align}
\label{Eq21}
\mathit{SSIM(x, y)=\frac{\left(2 \mu_{x} \mu_{y}+C_{1}\right)\left(2 \sigma_{x y}+C_{2}\right)}{\left(\mu_{x}^{2}+\mu_{y}^{2}+C_{1}\right)\left(\sigma_{x}^{2}+\sigma_{y}^{2}+C_{2}\right)}}
\end{align}
where $\mu $ and $\sigma$ denote the mean and standard deviation of an image. C1 and C2 are constants used to maintain the stability. The $L_{ssim}^{}$ can be written as Eq\eqref{Eq22}.
\begin{align}
\label{Eq22}
\mathit{L_{ssim}\left(J_{\mathrm{c}}, \hat{J}_{\mathrm{c}}\right)=1-\operatorname{SSIM}\left(J_{\mathrm{c}}, \hat{J}_{\mathrm{c}}\right)}
\end{align}

The $L_per$ is the perceptual loss\cite{johnson2016perceptual}. The perceptual loss function is expressed as:
\begin{align}
\label{Eq23}
\mathit{L_{p e r}\left(J_{\mathrm{c}}, \hat{J}_{\mathrm{c}}\right)=\left|\left(V G G\left(J_{\mathrm{c}}\right)-V G G\left(\hat{J}_{\mathrm{c}}\right)\right)\right|}
\end{align}
where $VGG$ refers to the $VGG-19$ network. The enhanced image and the ground truth image are passed to the last convolutional layer of the pretrained VGG network to get the feature representations. $\left | \cdot\right | $is the $L1$ distance. Integrating $L_{cha}^{}$,$L_{ssim}^{}$ and $L_{per}^{}$, the final loss function is written as:
\begin{align}
\label{Eq24}
\mathit{L_{ Total }=\lambda_{1} L_{\mathrm{cha}}+\lambda_{2} L_{ssim}+\lambda_{3} L_{   per }}
\end{align}
where $\lambda_1$, $\lambda_2$ and $\lambda_3$ denote the coefficients that adjust the importance of each loss component. During the training period, the values are set to 1,1.1,11 adjusted as a hyperparameter.

% \usepackage{multirow}
\begin{table*}[]
\centering
    \caption{Full reference image quality was evaluated with PSNR on the EUVP, UFO-120 and UIEB test sets.}
\begin{tabular}{cccccccccc}
\hline
\multicolumn{1}{l}{} & \multicolumn{1}{l}{} & \multicolumn{1}{l}{} & \multicolumn{1}{l}{} & \multicolumn{1}{l}{} & \multicolumn{1}{l}{} & \multicolumn{1}{l}{} & \multicolumn{1}{l}{} & \multicolumn{1}{l}{} & \multicolumn{1}{l}{} \\
Mtriy & Datesets & Cycle-GAN & DCP & Mul-Fusion & Water-Net & HE & FUnIE-GAN & UWNet & Ours \\
 &  &  &  &  &  &  &  &  &  \\ \hline
 &  &  &  &  &  &  &  &  &  \\
 & EUVP & 20.2253 & 18.7900 & 14.6648 & 23.8756 & 15.2762 & 23.4078 & 22.5752 & \textbf{25.2241} \\
 &  &  &  &  &  &  &  &  &  \\
PSNR$\uparrow$ & UFO-120 & 21.4797 & 19.5469 & 14.1857 & 24.4047 & 15.3250 & 23.9164 & 23.9877 & \textbf{26.3342} \\
 &  &  &  &  &  &  &  &  &  \\
 & UIEB & 19.3824 & 17.5542 & 15.4308 & 19.1130 & 17.4823 & 15.5455 & 18.7026 & \textbf{21.4253} \\
 &  &  &  &  &  &  &  &  &  \\ \hline
\end{tabular}
\end{table*}
% \usepackage{multirow}
\begin{table*}[]
\centering
    \caption{Full reference image quality was evaluated with SSIM on the EUVP, UFO-120 and UIEB test sets.}
\begin{tabular}{cccccccccc}
\hline
\multicolumn{1}{l}{} & \multicolumn{1}{l}{} & \multicolumn{1}{l}{} & \multicolumn{1}{l}{} & \multicolumn{1}{l}{} & \multicolumn{1}{l}{} & \multicolumn{1}{l}{} & \multicolumn{1}{l}{} & \multicolumn{1}{l}{} & \multicolumn{1}{l}{} \\
Mtriy & Datesets & Cycle-GAN & DCP & Mul-Fusion & Water-Net & HE & FUnIE-GAN & UWNet & Ours \\
 &  &  &  &  &  &  &  &  &  \\ \hline
 &  &  &  &  &  &  &  &  &  \\
 & EUVP & 0.6853 & 0.7525 & 0.5594 & 0.8129 & 0.6194 & 0.7916 & 0.7899 & \textbf{0.8265} \\
 &  &  &  &  &  &  &  &  &  \\
SSIM$\uparrow$ & UFO-120 & 0.8112 & 0.8209 & 0.6663 & 0.8652 & 0.6999 & 0.7565 & 0.7796 & \textbf{0.8819} \\
 &  &  &  &  &  &  &  &  &  \\
 & UIEB & 0.6549 & 0.6934 & 0.6716 & 0.7971 & 0.6348 & 0.5979 & 0.6646 & \textbf{0.8012} \\
 &  &  &  &  &  &  &  &  &  \\ \hline
\end{tabular}
\end{table*}
%%%%%%%%%%%%%%%%%%%%%%%%%%%%%%%%%%%%%%%%%%%%%%%%%%%%

\section{Experimental results}
HIFI-Net is compared with several SOTA (state-of-the-art) algorithms for experiments on three public available datasets. These three datasets are UFO-120\cite{islam2020simultaneous}, EUVP\cite{islam2020fast}, and UIEB\cite{li2019underwater}, where UFO-120 has a training set of 1620 pairs and a test set of 120 pairs, EUVP has a training set of 4535 pairs and a test set of 613 pairs, and UIEB has a training set of 800 pairs and a test set of 90 pairs. These experimental results involve several commonly used evaluation metrics (MSE, PSNR, SSIM) and a new proposed ER3C (The error of the ratio of three channels). ER3C is an evaluation metric to evaluate restoration ability of RGB structure . Then, the subjective evaluation experiments use some images from three datasets to show the enhancement of algorithms on underwater images. Finally, several ablation experiments are used to demonstrate the effectiveness of each module in HIFI-Net.

\subsection{commonly-used metrics}
MSE, PSNR, and SSIM are commonly employed for quantitative evaluations. MSE and PSNR are used to evaluate the restoration ability of algorithms in terms of image content. SSIM is used to evaluate the restoration ability of algorithms in terms of image structure and texture. The smaller the value of MSE, the stronger the algorithm. The larger the values of PSNR and SSIM, the stronger the algorithm.

As shown in Tab.I, the proposed HIFI-Net gets the lowest MSE meaning the best performance among all algorithms in three datasets. It is 26.7\% lower than the suboptimal algorithm Water-net on EUVP, 35.9\% lower than the suboptimal algorithm Water-net on UFO-120 and 52.7\% lower than the suboptimal algorithm Cyclegan on UIEB.

As shown in Tab.II, the HIFI-Net algorithm has the highest PSNR meaning the best performance among all algorithms in three datasets. It is 5.6\% higher than the suboptimal algorithm Water-net on EUVP, 7.3\% higher than the suboptimal algorithm Water-net on UFO-120 and 16.8\% higher than the suboptimal algorithm Cyclegan on UIEB.

As shown in Tab.III, the HIFI-Net algorithm has the highest SSIM meaning the best performance among all algorithms in three datasets. It is 1.7\% higher than the suboptimal algorithm Water-net on EUVP, 1.9\% higher than the suboptimal algorithm Water-net on UFO-120 and 0.9\% higher than the suboptimal algorithm Water-net on UIEB.

\subsection{The ER3C metric.}
The values of a pixel in an RGB image can be seen as the ratio, which means the RGB structure, and the average gray value. The scattering of the water causes the RGB structure be destroyed, therefore the color of an underwater object looks different from that in the air. This RGB degradation mathematically is a disruption to the ratio of three channel values. The underwater image enhancement algorithm should have RGB structure restoration ability. In this section, ER3C is proposed to evaluate the RGB structure restoration ability of the algorithms.
%%%%%%%%%%%%%%%%%%%%%%%%%%%%%%%%%%%%
\begin{table*}[]
\centering
    \caption{Full reference image quality was evaluated with ER3C on the EUVP, UFO-120 and UIEB test sets.}
\begin{tabular}{cccccccccc}
\hline
\multicolumn{1}{l}{} & \multicolumn{1}{l}{} & \multicolumn{1}{l}{} & \multicolumn{1}{l}{} & \multicolumn{1}{l}{} & \multicolumn{1}{l}{} & \multicolumn{1}{l}{} & \multicolumn{1}{l}{} & \multicolumn{1}{l}{} & \multicolumn{1}{l}{} \\
Mtriy & Datesets & Cycle-GAN & DCP & Mul-Fusion & Water-Net & HE & FUnIE-GAN & UWNet & Ours \\
 &  &  &  &  &  &  &  &  &  \\ \hline
 &  &  &  &  &  &  &  &  &  \\
 & EUVP & 5.2791 & 5.5019 & 6.1032 & 3.4509 & 8.4905 & 3.0251 & 6.6403 & \textbf{2.6190} \\
ER3C$\downarrow$ &  &  &  &  &  &  &  &  &  \\
$\left ( \times 10_{}^{-6} \right )$ &UFO-120 & 3.0739 & 5.1852 & 6.3273 & 3.0143 & 8.2079 & 2.8869 & 8.7498 & \textbf{2.2918} \\
 &  &  &  &  &  &  &  &  &  \\
 & UIEB & 25.1984 & 26.8712 & 26.6906 & 20.5563 & 31.6845 & 80.3736 & 68.1454 & \textbf{17.9162} \\
 &  &  &  &  &  &  &  &  &  \\ \hline
\end{tabular}
\end{table*}

%%%%%%%%%%%%%%%%%%%%%%%%%
\begin{table*}[]
\label{Tab.5}
\centering
    \caption{Numerical results of HIFI-Net structure ablation experiments on UIEB}
\begin{tabular}{llllllllllllllll}
\hline
\multicolumn{1}{c}{} & \multicolumn{1}{c}{} & \multicolumn{1}{c}{} & \multicolumn{1}{c}{} & \multicolumn{1}{c}{} & \multicolumn{1}{c}{} & \multicolumn{1}{c}{} & \multicolumn{1}{c}{} & \multicolumn{1}{c}{} & \multicolumn{1}{c}{} & \multicolumn{1}{c}{} & \multicolumn{1}{c}{} & \multicolumn{1}{c}{} & \multicolumn{1}{c}{} & \multicolumn{1}{c}{} & \multicolumn{1}{c}{} \\
\multicolumn{2}{c}{} & \multicolumn{2}{c}{Haar} & \multicolumn{2}{c}{CBAM} & \multicolumn{2}{c}{RFM-Haar} & \multicolumn{2}{c}{Convs} & \multicolumn{3}{c}{PSNR$\uparrow$} & \multicolumn{3}{c}{SSIM$\uparrow$} \\
\multicolumn{1}{c}{} & \multicolumn{1}{c}{} & \multicolumn{1}{c}{} & \multicolumn{1}{c}{} & \multicolumn{1}{c}{} & \multicolumn{1}{c}{} & \multicolumn{1}{c}{} & \multicolumn{1}{c}{} & \multicolumn{1}{c}{} & \multicolumn{1}{c}{} & \multicolumn{1}{c}{} & \multicolumn{1}{c}{} & \multicolumn{1}{c}{} & \multicolumn{1}{c}{} & \multicolumn{1}{c}{} & \multicolumn{1}{c}{} \\ \hline
 &  &  &  &  &  &  &  &  &  &  &  &  &  &  &  \\
\multicolumn{1}{c}{Exp1} & \multicolumn{1}{c}{} & \multicolumn{1}{c}{\checkmark} & \multicolumn{1}{c}{} & \multicolumn{1}{c}{} & \multicolumn{1}{c}{} & \multicolumn{1}{c}{\checkmark} & \multicolumn{1}{c}{} & \multicolumn{1}{c}{} & \multicolumn{1}{c}{} & \multicolumn{3}{c}{18.82$\pm$3.048} & \multicolumn{3}{c}{0.770$\pm$0.081} \\
 &  &  &  &  &  &  &  &  &  &  &  &  &  &  &  \\
 &  &  &  &  &  &  &  &  &  &  &  &  &  &  &  \\
\multicolumn{1}{c}{Exp2} & \multicolumn{1}{c}{} & \multicolumn{1}{c}{} & \multicolumn{1}{c}{} & \multicolumn{1}{c}{\checkmark} & \multicolumn{1}{c}{} & \multicolumn{1}{c}{} & \multicolumn{1}{c}{} & \multicolumn{1}{c}{\checkmark} & \multicolumn{1}{c}{} & \multicolumn{3}{c}{20.54$\pm$3.048} & \multicolumn{3}{c}{0.781$\pm$0.0744} \\
 &  &  &  &  &  &  &  &  &  &  &  &  &  &  &  \\
 &  &  &  &  &  &  &  &  &  &  &  &  &  &  &  \\
\multicolumn{1}{c}{Exp3} & \multicolumn{1}{c}{} & \multicolumn{1}{c}{\checkmark} & \multicolumn{1}{c}{} & \multicolumn{1}{c}{\checkmark} & \multicolumn{1}{c}{} & \multicolumn{1}{c}{} & \multicolumn{1}{c}{} & \multicolumn{1}{c}{\checkmark} & \multicolumn{1}{c}{} & \multicolumn{3}{c}{20.22$\pm$0.2.892} & \multicolumn{3}{c}{0.771$\pm$0.080} \\
 &  &  &  &  &  &  &  &  &  &  &  &  &  &  &  \\
 &  &  &  &  &  &  &  &  &  &  &  &  &  &  &  \\
\multicolumn{1}{c}{Exp4} & \multicolumn{1}{c}{} & \multicolumn{1}{c}{\checkmark} & \multicolumn{1}{c}{} & \multicolumn{1}{c}{\checkmark} & \multicolumn{1}{c}{} & \multicolumn{1}{c}{\checkmark} & \multicolumn{1}{c}{} & \multicolumn{1}{c}{} & \multicolumn{1}{c}{} & \multicolumn{3}{c}{21.42$\pm$2.638} & \multicolumn{3}{c}{0.801$\pm$0.062} \\
 &  &  &  &  &  &  &  &  &  &  &  &  &  &  &  \\ \hline
\end{tabular}
\end{table*}
%%%%%%%%%%%%%%%%%%%%%%%%%%%%%%%%%%%%%%%%%%%%%%%%%%%%
Assuming the image is $I$, where $\mathit{I(x,y)}$ is a pixel at the position (x, y). The pixel values on three channel are $\mathit{I(x, y, 1)}$, $\mathit{I(x, y, 2)}$, $\mathit{I(x, y, 3)}$ respectively.  $\hat{I}$ is the ground truth image. 
\begin{align}
\label{Eq24}
\mathit{i_{a}(x, y)=\sum_{z=1}^{3} I(x, y, z) / 3} \quad\quad\quad\quad \\  
\mathit{\hat{i}_{a}(x, y)=\sum_{z=1}^{3} \hat{I}(x, y, z) / 3} \quad\quad\quad\quad\\
\mathit{E R 3 C=(1 / M N) \sum_{x=1}^{M} (\sum_{y=1}^{N}(\sum_{z=1}^{3}}\quad\quad \nonumber\\
\mathit{|I(x, y, z) / i_{a}(x, \mathrm{y})-\hat{I}(x, y, z) / \hat{i}_{a}(x, y)|))} 
\end{align}
where $\mathit{i_a(x, y)}$ is the average gray value of $\mathit{I(x,y)}$ pixels and $\hat{i}_{a}(x, y)$ is the average gray value of $\mathit{\hat{I}(x, y)}$  pixels, in the three channels. 

As shown in Tab.IV, the HIFI-Net algorithm has the lowest ER3C meaning the best performance among all algorithms on all three datasets. The proposed HIFI-Net is 24.1\% lower than the suboptimal algorithm WaterNet In EUVP, 24.0\% lower than the suboptimal algorithm WaterNet in UFO-120 and 12.8\% lower than the suboptimal algorithm WaterNet in the UIEB. The proposed HiFI-Net performs better than other algorithms in terms of RGB structure restoration ability.

\subsection{Subjective evaluation}
As the Fig.4 shows, this section shows the enhancement of multiple algorithms on three datasets. The first two rows are on the EUVP dataset. The third and fourth rows are on the UFO-120 dataset. The fifth and sixth rows are on the UIEB dataset. It can be seen that HIFI-Net is better in color restoration ability, compared to other algorithms.

%%%%%%%%%%%%%%%%%%%%%%%%%%%%%%%%%%%%%%%%%%%%%
\begin{figure*}[!t]
\centering
\includegraphics[width=2.0\columnwidth,height =1.3\columnwidth]{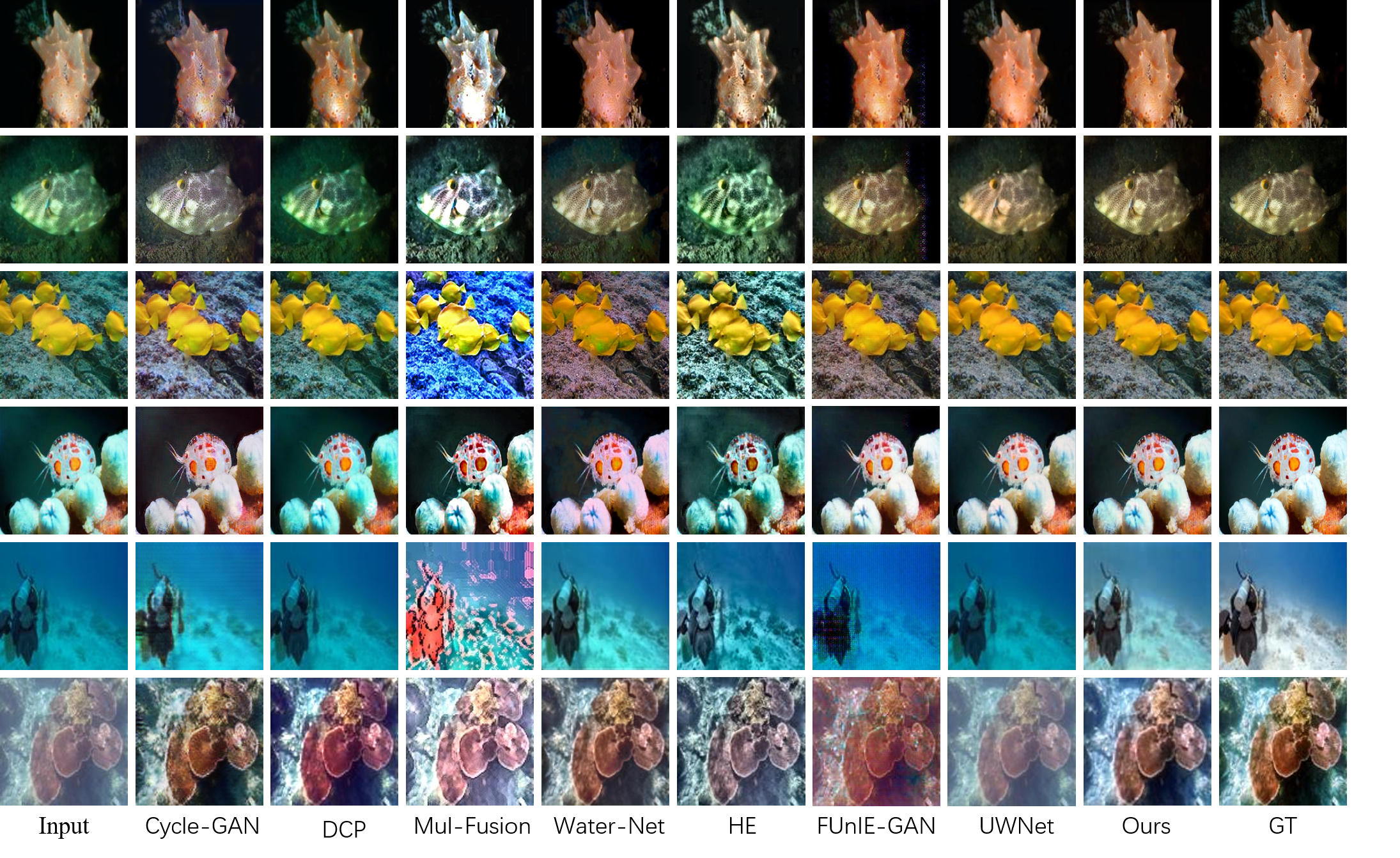}
\caption{Columns from left to right show input underwater images and the results of Cycle-GAN,
DCP, Mul-Fusion, Water-Net, HE, FUnIE-GAN, UWNet, the proposed HIFI-Net. The last column shows GT (ground truth) images.}
\label{Fig4}
\end{figure*}
%%%%%%%%%%%%%%%%%%%%%%%%%%%%%%%%%%%%%%%%%%%%%

\subsection{Ablation experiments}
To further verify the effectiveness of the HIFI-Net algorithm, ablation experiments are conducted on the key modules. The verifying experiments mainly include: the effect of Haar wavelet transform; the effectiveness of the RFM-Haar; the effectiveness of CBAM on the algorithm. The Convs in Tab.V indicates the convolution block composed of six convolutions of different kernels. This convolution block is used to fuse the original image and Haar wavelet transform results without RFM.

In Exp3 in Tab.V, the underwater images are pre-processed using Haar wavelet transform. The pre-processing results are input into the network. Exp3 and Exp4 are conducted to verify the effectiveness of the RFM-Haar. In Exp3, Convs is used instead of RFM to ensure the fairness. It can be seen that simply combining Haar wavelet transform with CNN to enhance underwater images is difficult to obtain better results compared to with RFM-Haar, as the PSNR and SSIM values in Exp3 is worse than in Exp4. RFM-Haar results in an improvement in terms of enhancement ability and robustness.

In Exp2 in Tab.V, Haar wavelet and RFM are not used. Within Exp2 and Exp3, It can be seen that simply combining Haar can even lead to degradation of CNN, as the PSNR and SSIM values in Exp3 is worse than in Exp2.

Exp1 and Exp4 in Tab.V are conducted to verify the effectiveness of CBAM module on the network. In HIFI-Net, CBAM is used as a supplementary to the RFM-Haar. Exp4 shows improvement compared to Exp1. It can be seen that using RFM-Haar in combination with CBAM in underwater image enhancement is a reasonable choice.

%%%%%%%%%%%%%%%%%%%%%%%%%%%%%%%%%%%%%%%%
\begin{table*}[]
\centering
    \caption{Numerical results of RFM-Haar structure ablation experiments on UIEB}
\begin{tabular}{ccccccccccccll}
\hline
\multicolumn{1}{l}{} & \multicolumn{1}{l}{} & \multicolumn{1}{l}{} & \multicolumn{1}{l}{} & \multicolumn{1}{l}{} & \multicolumn{1}{l}{} & \multicolumn{1}{l}{} & \multicolumn{1}{l}{} & \multicolumn{1}{l}{} & \multicolumn{1}{l}{} & \multicolumn{1}{l}{} & \multicolumn{1}{l}{} &  &  \\
 &  & w/o Residual &  & w/o Maxpool &  &  w/o MLP &  & w/o Base image &  & \multicolumn{4}{c}{Ours} \\
\multicolumn{2}{c}{} & \multicolumn{2}{c}{} & \multicolumn{2}{c}{} & \multicolumn{2}{c}{} & \multicolumn{2}{c}{} & \multicolumn{2}{c}{} &  &  \\ \hline
\multicolumn{1}{l}{} & \multicolumn{1}{l}{} & \multicolumn{1}{l}{} & \multicolumn{1}{l}{} & \multicolumn{1}{l}{} & \multicolumn{1}{l}{} & \multicolumn{1}{l}{} & \multicolumn{1}{l}{} & \multicolumn{1}{l}{} & \multicolumn{1}{l}{} & \multicolumn{1}{l}{} & \multicolumn{1}{l}{} &  &  \\
\multicolumn{2}{c}{PSNR$\uparrow$} & 20.5612 &  & 18.9985 &  & 18.8140 &  & 19.3252 &  & \multicolumn{4}{c}{21.42} \\
 &  &  &  &  &  &  &  &  &  &  &  &  &  \\
 &  &  &  &  &  &  &  &  &  &  &  &  &  \\
\multicolumn{2}{c}{SSIM$\uparrow$} & 0.7809 &  & 0.7716 &  & 0.7731 &  & 0.7822 &  & \multicolumn{4}{c}{0.8012} \\
 &  &  &  &  &  &  &  &  &  &  &  &  &  \\ \hline
\end{tabular}
\end{table*}
%%%%%%%%%%%%%%%%%%%%%%%%%%%%%%%%%%%%%%%%%%

Tab.VI shows the ablation experiments on the components within the RFU and the residual blocks used to connect the RFU. It can be seen that both metrics of the network decreased after removing residual or any component in RFU.

\bibliographystyle{ieeetr}
\bibliography{paper}

\begin{thebibliography}{10}

\bibitem{woo2018cbam}
S.~Woo, J.~Park, J.-Y. Lee, and I.~S. Kweon, ``Cbam: Convolutional block
  attention module,'' in {\em Proceedings of the European conference on
  computer vision (ECCV)}, pp.~3--19, 2018.

\bibitem{li2018emerging}
C.~Li, J.~Guo, and C.~Guo, ``Emerging from water: Underwater image color
  correction based on weakly supervised color transfer,'' {\em IEEE Signal
  processing letters}, vol.~25, no.~3, pp.~323--327, 2018.

\bibitem{he2010single}
K.~He, J.~Sun, and X.~Tang, ``Single image haze removal using dark channel
  prior,'' {\em IEEE transactions on pattern analysis and machine
  intelligence}, vol.~33, no.~12, pp.~2341--2353, 2010.

\bibitem{mohan2020underwater}
S.~Mohan and P.~Simon, ``Underwater image enhancement based on histogram
  manipulation and multiscale fusion,'' {\em Procedia Computer Science},
  vol.~171, pp.~941--950, 2020.

\bibitem{li2019underwater}
C.~Li, C.~Guo, W.~Ren, R.~Cong, J.~Hou, S.~Kwong, and D.~Tao, ``An underwater
  image enhancement benchmark dataset and beyond,'' {\em IEEE Transactions on
  Image Processing}, vol.~29, pp.~4376--4389, 2019.

\bibitem{pizer1987adaptive}
S.~M. Pizer, E.~P. Amburn, J.~D. Austin, R.~Cromartie, A.~Geselowitz, T.~Greer,
  B.~ter Haar~Romeny, J.~B. Zimmerman, and K.~Zuiderveld, ``Adaptive histogram
  equalization and its variations,'' {\em Computer vision, graphics, and image
  processing}, vol.~39, no.~3, pp.~355--368, 1987.

\bibitem{islam2020fast}
M.~J. Islam, Y.~Xia, and J.~Sattar, ``Fast underwater image enhancement for
  improved visual perception,'' {\em IEEE Robotics and Automation Letters},
  vol.~5, no.~2, pp.~3227--3234, 2020.

\bibitem{naik2021shallow}
A.~Naik, A.~Swarnakar, and K.~Mittal, ``Shallow-uwnet: Compressed model for
  underwater image enhancement (student abstract),'' in {\em Proceedings of the
  AAAI Conference on Artificial Intelligence}, vol.~35, pp.~15853--15854, 2021.

\bibitem{lai2018fast}
W.-S. Lai, J.-B. Huang, N.~Ahuja, and M.-H. Yang, ``Fast and accurate image
  super-resolution with deep laplacian pyramid networks,'' {\em IEEE
  transactions on pattern analysis and machine intelligence}, vol.~41, no.~11,
  pp.~2599--2613, 2018.

\bibitem{wang2004image}
Z.~Wang, A.~C. Bovik, H.~R. Sheikh, and E.~P. Simoncelli, ``Image quality
  assessment: from error visibility to structural similarity,'' {\em IEEE
  transactions on image processing}, vol.~13, no.~4, pp.~600--612, 2004.

\bibitem{johnson2016perceptual}
J.~Johnson, A.~Alahi, and L.~Fei-Fei, ``Perceptual losses for real-time style
  transfer and super-resolution,'' in {\em European conference on computer
  vision}, pp.~694--711, Springer, 2016.

\bibitem{islam2020simultaneous}
M.~J. Islam, P.~Luo, and J.~Sattar, ``Simultaneous enhancement and
  super-resolution of underwater imagery for improved visual perception,'' {\em
  arXiv preprint arXiv:2002.01155}, 2020.

\end{thebibliography}
\end{document}